\documentclass[10pt,twocolumn,letterpaper]{article}

\usepackage{iccv}
\usepackage{times}
\usepackage{epsfig}
\usepackage{graphicx}
\usepackage{amsmath}
\usepackage{amssymb}
\usepackage{booktabs}

\usepackage[breaklinks=true,bookmarks=false]{hyperref}

\iccvfinalcopy 

\def\iccvPaperID{****} 

\ificcvfinal\fi

\begin{document}

\title{SRPCN: Structure Retrieval based Point Completion Network}

\author{Kaiyi Zhang, Ximing Yang, Yuan Wu, Cheng Jin\\
School of Computer Science, Fudan University, Shanghai, China\\
{\tt\small \{zhangky20, xmyang19, wuyuan, jc\}@fudan.edu.cn}
}

\maketitle
\ificcvfinal\thispagestyle{empty}\fi

\begin{abstract}
   Given partial objects and some complete ones as references, point cloud completion aims to recover authentic shapes. 
   However, existing methods pay little attention to general shapes, 
   which leads to the poor authenticity of completion results. 
   Besides, the missing patterns are diverse in reality, but existing methods can only handle 
   fixed ones, which means a poor generalization ability. 
   Considering that a partial point cloud is a subset of the corresponding complete one, 
   we regard them as different samples of the same distribution 
   and propose Structure Retrieval based Point Completion Network (SRPCN). 
   It first uses k-means clustering to extract structure points and disperses them into distributions, 
   and then KL Divergence is used as a metric to find the complete structure point cloud 
   that best matches the input in a database. 
   Finally, a PCN-like decoder network is adopted to generate the final results 
   based on the retrieved structure point clouds. 
   As structure plays an important role in describing the general shape of an object
   and the proposed structure retrieval method is robust to missing patterns, 
   experiments show that our method can generate more authentic results and has a stronger generalization ability.
\end{abstract}

\section{Introduction}

Point cloud completion aims to complete authentic point clouds when given inputs with various missing patterns. 
It can contribute to a series of downstream applications, 
like robotics operations\cite{varley2017shape}, scene understanding\cite{dai2018scancomplete}, 
and virtual operations of complete shapes\cite{visapp17}.

\begin{figure}[t]
\begin{center}
   \includegraphics[width=1.0\linewidth]{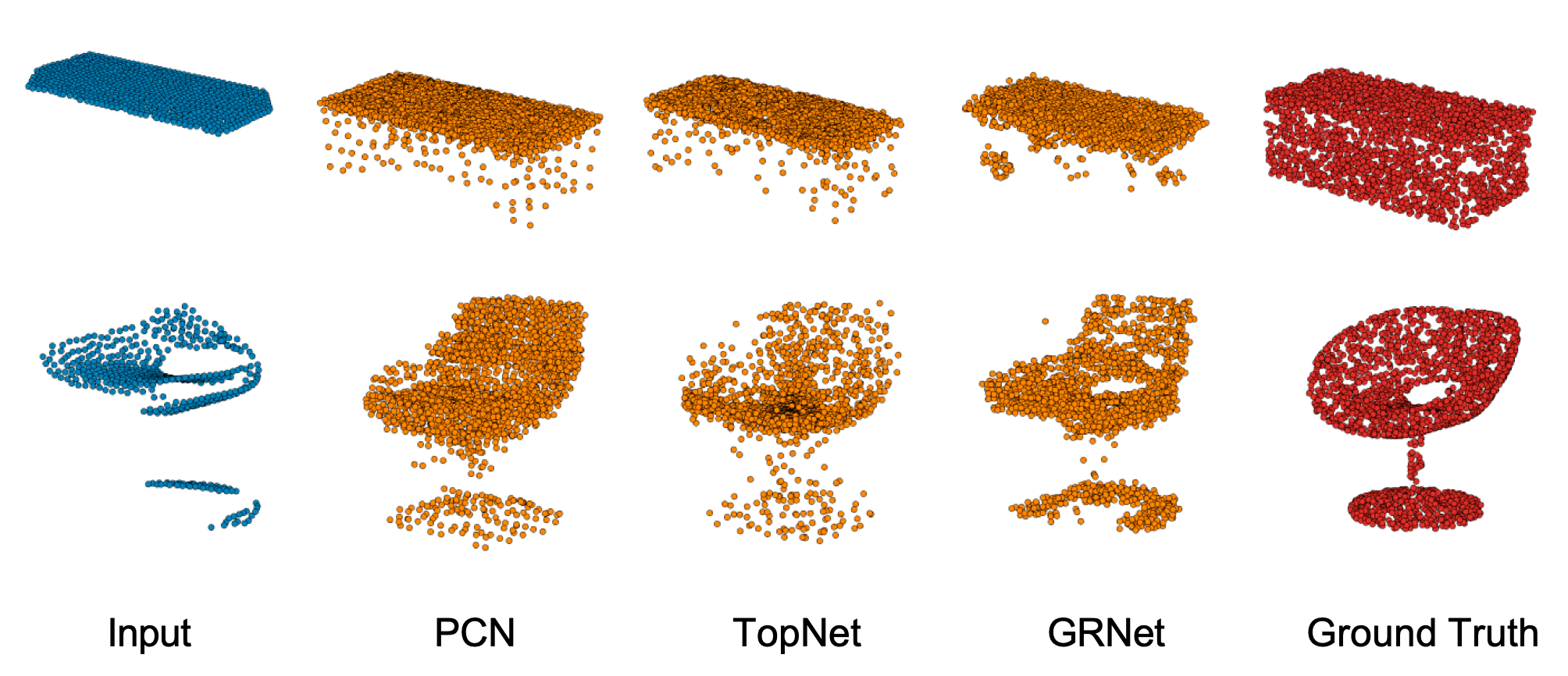}
\end{center}
   \caption{The completion results of a cabinet and a chair in PCN\cite{yuan2018pcn}, TopNet\cite{topnet2019} 
   and GRNet\cite{xie2020grnet}, which have fuzzy structures and poor authenticity.}
\label{fig:previous_weakness}
\end{figure}

Most of the recent point cloud completion methods are related with deep learning
and achieve promising results. However, they have two problems. 
First, the authenticity of the completion results is poor. 
As shown in Figure \ref{fig:previous_weakness}, 
we can find that the results in existing methods are fuzzy and unreasonable. 
This may be caused by two reasons: 
the fuzzy structure and the usage of Chamfer Distance\cite{Fan_2017_CVPR}. 
Structure plays an important role in describing the general shape of an object 
and helps to understand the object itself, but existing methods\cite{liu2019morphing, topnet2019, Wang_2020_CVPR, 
DBLP:journals/corr/abs-2008-07358, Wen_2020_CVPR, xie2020grnet, yuan2018pcn, zhang2020preserved}
do not explicitly consider this low-frequency information. Besides, 
many methods\cite{topnet2019, Wang_2020_CVPR, 
DBLP:journals/corr/abs-2008-07358, Wen_2020_CVPR, yuan2018pcn, zhang2020preserved} 
use Chamfer Distance as a loss function or an evaluation metric, 
but as mentioned by \cite{achlioptas2017latent_pc, liu2019morphing}, 
Chamfer Distance might be blind to some visual inferiority. 

Second, existing methods\cite{liu2019morphing, topnet2019, Wang_2020_CVPR, 
DBLP:journals/corr/abs-2008-07358, Wen_2020_CVPR, xie2020grnet, yuan2018pcn, zhang2020preserved}
fix missing patterns in advance, such as, back-projecting 2.5D depth images into 3D, 
removing some points within a certain radius from complete point clouds in a view-point. 
This leads to the poor generalization ability of their models. 
Although PF-Net\cite{huang2020pfnet} tries to tackle this situation, the new missing parts for testing are 
close to what in the training set and are only experimented in the airplane category. What's more, 
PF-Net makes a dataset based on ShapeNet-Part dataset\cite{Yi16} 
with a fixed missing rate, which rarely happens in real scenarios.

To address these issues, we propose Structure Retrieval based Point Completion Network. 
It mainly includes two steps: structure retrieval and retrieval recovery. 
Specifically, we first use k-means clustering to extract structure points and disperses them into distributions. 
Then KL Divergence is used as a metric to find the complete structure point cloud 
that best matches the input in a database. 
Finally, through adopting a PCN-like decoder network, 
we upsample the retrieved structure point clouds to obtain final results. 
Because of the reasonable structure point clouds obtained by structure retrieval, which represent general shapes, 
our method can achieve more authentic completion results. Besides, 
as the structure retrieval is robust to missing patterns, our method has a stronger generalization ability.

Our main contributions are the following:
\begin{itemize}
   \item We propose a new representation of the point cloud, structure distribution, 
   which well preserves the general shape with a smaller number of points.
   \item We continuous the structure point clouds into distributions 
   and propose a fast structure retrieval method based on KL Divergence.
   \item Our proposed method achieves more authentic completion results and 
   shows stronger generalization ability on the Completion3D dataset.
\end{itemize}

\section{Related work}

\begin{figure*}[h]
\begin{center}
    \includegraphics[width=1.0\linewidth]{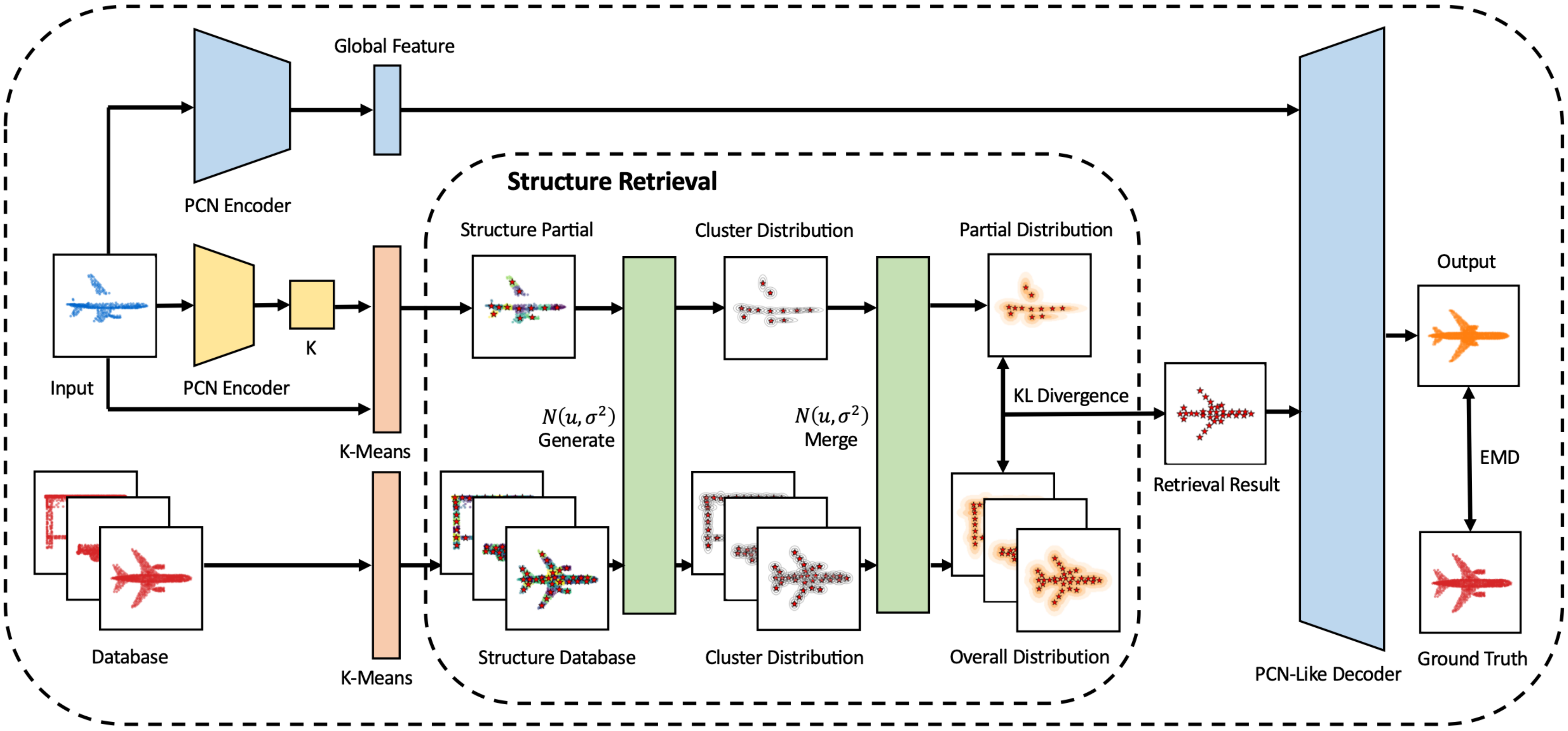}
\end{center}
   \caption{\textbf{The Architecture of SRPCN.} The three-color blocks of yellow, green, and blue represent different parts. 
   The most critical and novel part is Structure Retrieval and it can be divided into three steps: 
   K-means clustering is adopted to get structure points; each structure point forms an ellipsoidal Gaussian distribution; 
   these distributions are merged to form an overall distribution and KL Divergence is used to guide the retrieval.}
\label{fig:overall_framework}
\end{figure*}

\subsection{Alignment-based 3D Shape Completion}

Alignment is a commonly used method in 3D shape completion\cite{li2015database, 
Nan_asearch-classify, SGP:SGP05:023-032, Shao:2012:IAS:2366145.2366155, Shen:2012, Sung:2015}. 
It usually includes two steps: 
The first step is retrieving some similar template shapes, 
which can be complete or part of shapes from a large shape database.
The second step is deforming and assembling the matched shapes to 
finish the completion. 
For example, Shen et al.\cite{Shen:2012} assemble reasonably labeled parts to 
complete the low-quality scanned point cloud data, from a small-scale shape repository. 
Sung et al.\cite{Sung:2015} associate the local coordinate system with existing shape parts 
and learn the position and orientation distribution of all other parts from the database. 
This method uses shape databases for completion and contributes a lot to the 3D reconstruction task.

Since the shape databases are in mesh format, all the methods above 
finally get the mesh shape, not only the matching speed is slow 
but also the output data format is not what we need. 
In contrast, we build a database in the format of the point cloud. The database is based on the structure 
point clouds, so the speed of retrieval in our method is fast. 
Through combining deep learning methods for upsampling, final shapes completed by our method 
have strong authenticity.

\subsection{Deep Learning on Point Cloud Completion}

The method of deep learning for point cloud research is first proposed 
by PointNet\cite{qi2016pointnet} 
and achieves great success in 3d point cloud classification\cite{NEURIPS2018_f5f8590c, liu2019rscnn, 
qi2017pointnetplusplus, thomas2019KPConv, uy-scanobjectnn-iccv19, dgcnn, wu2018pointconv}. This causes 
the popularity of employing deep learning on the point cloud.
About 3D point cloud completion, 
existing deep learning methods can be roughly divided into two categories: 

\textbf{Directly output the complete shape}. 
This method does not need to know the missing parts and the missing 
rate of shapes, instead, are obtained implicitly through deep learning. 
However, due to the lack of attention to reasonable structures, the completion results are not very authentic. 
PCN\cite{yuan2018pcn} provides an autoencoder to combine 
the global and local shape information for point cloud completion. 
TopNet\cite{topnet2019} further designs a tree-structured decoder to realize multi-scale completion. 
MSN\cite{liu2019morphing} uses multi-MLPs as patches to generate better local shapes 
and states that using the metric of Earth's Mover Distance\cite{Fan_2017_CVPR} is more reasonable than 
Chamfer Distance\cite{Fan_2017_CVPR}. 
GRNet\cite{xie2020grnet} transfers point cloud into a new voxel representation, 
which better retains the spatial information of original partial point clouds. 
In addition, \cite{groueix2018, Sarmad_2019_CVPR, Wang_2020_CVPR, 
DBLP:journals/corr/abs-2008-07358, Wen_2020_CVPR, Yang_2018_CVPR, zhang2020preserved} 
play an important role in promoting the point cloud completion task.

\textbf{Complete the missing parts and then merge}.
This method well retains existing point clouds and only fills in the missing parts of shapes. 
But existing work assumes that 
the missing parts and the missing rate are known, which greatly reduces the difficulty of completion. 
PF-Net\cite{huang2020pfnet} provides a multi-resolution encoder to get better feature embeddings 
and a point pyramid decoder to complete the missing parts progressively, which is a heuristic work.

In this paper, we combine deep-learning-based methods with alignment-based methods. 
To make our method more reliable, 
we first use a PCN-like network to predict the missing rate. 
Based on this, we further design a structure retrieval method based on KL Divergence 
to predict structure point clouds of the missing parts. 
Finally, a PCN-like decoder is used for upsampling the retrieved complete structure point clouds. 
Even when the missing parts and the missing rate are indeterminate, our method still can complete
reasonable point clouds.

\section{Methods}

In this section, we describe the architecture of our model SRPCN. 
As shown in Figure \ref{fig:overall_framework}, 
the input of SRPCN includes partial point clouds and a database generated from 
complete point clouds in the trainset and the output of SRPCN is the completed 2048 point clouds.
SRPCN first determines the structure of point clouds 
through structure retrieval and then combines the PCN\cite{yuan2018pcn} network to achieve completion.
It includes three parts: Missing Rate Prediction, Structure Retrieval, and Point Cloud Upsampling. 
In the Missing Rate Prediction, we use the architecture of the encoder in PCN to predict 
the missing rate of the input, which determines the number of points 
in the partial structure point cloud. In the Structure Retrieval, 
we first use k-means clustering to extract structure points and disperses them into distributions, 
and then KL Divergence is used as a metric to find the complete structure point cloud 
that best matches the input in a database. In the Point Cloud Upsampling,
we use a PCN-like decoder to upsample the matched structure point clouds.
Next, we will describe the specific design of SRPCN from the three parts.

\subsection{Missing Rate Prediction}

In reality, the missing rates of partial point clouds are often uncertain. For example, 
the partial point clouds in the Completion3D\cite{topnet2019} dataset are generated 
by back-projecting 2.5D depth images into 3D and thus different views lead to 
different missing rates. Since the distribution matching in our structure retrieval 
also contains structure information, 
it is necessary to extract structure points from the input while keeping the missing rate basically unchanged. 
To this end, we adopt a neural network to predict the missing rate. 
We follow the architecture of the PCN Encoder while adding a fully connected layer 
($1024 \rightarrow 256 \rightarrow 64 \rightarrow 1$) to design the network.
This part requires pre-training and the L1 loss function is used for optimizing.

\subsection{Structure Retrieval}

The main purpose of this part is to realize the prediction of the missing parts. 
Inspired by the retrieval methods used in traditional point cloud reconstruction, 
we try to borrow this method to point cloud completion. 
However, if we suppose the retrieval method is based on complete point cloud data, 
the time and space complexity will be very high. 
For this, we first downsample point clouds 
into structure point clouds and then do retrieval on this small scale, 
which greatly accelerates the efficiency of retrieval. 
Our proposed structure retrieval method can be divided into the following three steps: 
K-means clustering is adopted to get structure points; each structure point forms an ellipsoidal Gaussian distribution; 
these distributions are merged to form an overall distribution and KL Divergence is used to guide the retrieval. 
With the retrieved structures, 
the authenticity of final completion results is greatly improved.

\subsubsection{Definition of KL Divergence}

Kullback-Leibler Divergence (KLD), also called relative entropy in the information system, 
randomness in continuous time series, and information gain in statistical model inference, 
is proposed by Kullback et al.\cite{10.1214/aoms/1177729694}. 
KL Divergence is a measure of the asymmetry of the difference 
between two probability distributions $P$ and $Q$. The KL Divergence is a measure of the average 
number of extra bits required to use the Q-based distribution to encode samples that 
follow the $P$ distribution. Typically, $P$ represents the true distribution of the data, 
and $Q$ represents the theoretical distribution of the data, the estimated model distribution, 
or the approximate distribution of $P$.

For discrete random variables, the KL Divergence of the probability distributions $P$ and $Q$ can be defined as:
\begin{equation}
D_{KL}(P||Q) = \sum_{x} P(x) ln \frac{P(x)}{Q(x)}
\end{equation}

For continuous random variables, the KL Divergence of the probability distributions $P$ and $Q$ can be defined as: 
\begin{equation}
D_{KL}(P||Q) = \int_{-\infty}^{\infty} p(x) ln \frac{p(x)}{q(x)}
\end{equation}

Through the definition and formula of KL Divergence, we can find that it is 
unidirectional and asymmetric. If we regard distribution $P$ as a distribution of a 
partial point cloud and $Q$ as a distribution of the corresponding complete point cloud, 
$D_{KL}(P||Q)$ will measure the degree 
of matching between the partial point cloud and the complete point cloud, 
so the value of KL Divergence can be used to guide the structure retrieval.

\subsubsection{Unilateral Discrete}

Point cloud data is discrete and there is no good distribution to describe it currently. 
Besides, it is difficult for the computer to calculate 
the KL Divergence of the two distributions $P$ and $Q$ with complicated formulas. 
Therefore, the KL Divergence of continuous random variables may not be suitable for point cloud completion. 
Another idea is to match partial point clouds and complete point clouds without downsampling 
by using discrete KL Divergence, but we have stated before that the time and space complexity will be very high. 
Therefore, it is a reasonable solution to downsample the point clouds to structure points 
and then perform matching. Since the commonly used downsampling methods, such as 
farthest point sampling (FPS) and Poisson disk sampling (PDS)\cite{Wei08parallelpoisson}, 
are unstable and cannot well represent the structure of point clouds, 
we finally use k-means clustering to extract structure point clouds. 
Because of the instability of k-means clustering, we diffuse complete structure point clouds 
into distributions to realize a unilateral discrete KL Divergence matching.

The distribution formation of complete structure point clouds in the database 
is shown in the Structure Retrieval part of Figure \ref{fig:overall_framework}. 
There are three steps. 
First, we perform k-means clustering on partial point clouds and complete point clouds. 
Each partial point cloud forms $(1-missing \; rate) \times K$ clusters, and 
each complete point cloud forms $K$ clusters. The center of each cluster is 
treated as a structure point. As shown by the Structure Partial or the Structure Database in Figure \ref{fig:overall_framework}, 
different colors indicate different clusters, and the $\star$ means the center of a cluster. 
Second, we take the variance of each cluster in the three dimensions $(x, y, z)$ 
as the variance of the corresponding structure point. 
With the center point $\mu$ and the variance $\sigma^2$, each structure point can form 
a Gaussian distribution $N(\mu, \sigma^2)$. As shown by the Cluster Distribution in Figure \ref{fig:overall_framework}, 
each structure point is diffused to form an ellipsoidal Gaussian distribution. 
Finally, we accumulate these Gaussian distributions corresponding to all the structure points to form the overall distribution.

With the description of the distribution, 
we can measure the degree of matching between two structure point clouds. 
Here we set $DB$ as the database, $Y_X$ as the target of the retrieval, 
$X$ as the partial structure point cloud and the corresponding distribution is $P_X$,
$Y$ as the complete structure point cloud in the database and the corresponding distribution is $Q_Y$. 
In the unilateral discrete, $P_X$ is a fixed discrete value, and thus the goal of structure retrieval is:
\begin{equation}
\begin{aligned}
Y_X 
&= \mathop{\arg\!\min}_{Y \in DB} D_{KL} (P_X || Q_Y) \\
&\Leftrightarrow \mathop{\arg\!\min}_{Y \in DB} \sum_{x \in X} ln \frac{1}{Q_Y(x)} \\
&\Leftrightarrow \mathop{\arg\!\max}_{Y \in DB} \sum_{x \in X} Q_Y(x) \\
&\Leftrightarrow \mathop{\arg\!\max}_{Y \in DB} \sum_{x \in X} \sum_{\mu \in Y} G(x; \mu, \sigma'^2)
\end{aligned}
\end{equation}

Since the three dimensions $(x, y, z)$ are not related, Gaussian distribution can be written as:
\begin{equation}
G(x; \mu, \sigma^2) = \prod_{i=1}^3 \frac{1}{\sqrt{2\pi} \sigma_i} exp(-\frac{(x_i - \mu_i)^2}{2\sigma_i^2})
\end{equation}

Considering that different objects have different structure point cloud densities 
(for example, the structure point cloud density of a lamp 
is usually greater than that of a car), 
the value range of the Gaussian distribution is different, 
which is very important for cross-category structure retrieval. 
Therefore, we first normalize the Gaussian distribution of each structure point 
to achieve the same value range of the final distributions in different objects. 
The normalization process is as follows:
\begin{equation}
\begin{aligned}
\begin{cases}
\begin{aligned}
\sigma_1' \sigma_2' \sigma_3' &= (\frac{\lambda}{\sqrt{2\pi}})^3 \\
\sigma_1':\sigma_2':\sigma_3' &= \sigma_1:\sigma_2:\sigma_3 \\
\end{aligned}
\end{cases} \\
\Rightarrow
\begin{cases}
\begin{aligned}
\sigma_1'^2 &= \frac{\lambda^2}{2\pi} (\frac{\sigma_1^2}{\sigma_2\sigma_3})^{\frac{2}{3}} \\ 
\sigma_2'^2 &= \frac{\lambda^2}{2\pi} (\frac{\sigma_2^2}{\sigma_1\sigma_3})^{\frac{2}{3}} \\ 
\sigma_3'^2 &= \frac{\lambda^2}{2\pi} (\frac{\sigma_3^2}{\sigma_1\sigma_2})^{\frac{2}{3}} \\
\end{aligned}
\end{cases}
\end{aligned}
\end{equation}

Here $\lambda$ is a scaling ratio suitable for the dataset. 
We have the final Gaussian distribution function for each structure point as follows:
\begin{equation}
G(x; \mu, \sigma'^2) = \prod_{i=1}^3 \frac{1}{\lambda} exp(-\frac{(x_i - \mu_i)^2}{2\sigma_i'^2})
\end{equation}

Through retrieving in a structure point cloud database, we will get the complete structure point cloud with 
the smallest KL Divergence value. 
In order to enhance the rationality and accuracy of the retrieval, we further optimize the formula. 
We consider that the partial point cloud is a 
subset of the complete point cloud, so the partial structure point cloud should 
completely fall into the distribution formed by the complete structure point cloud. 
Therefore, we design a threshold $\gamma$. When a certain point of the partial 
structure point cloud, its calculated value in the distribution is less than the threshold, 
we believe that they are not a match. 
The improved formula is as follows:
\begin{equation}
Y_X = \mathop{\arg\!\max}_{Y \in DB}
\begin{cases}
0,              \text{$\exists_{x \in X} \sum_{\mu \in Y} G(x; \mu, \sigma'^2) \le \gamma$} \\
\sum_{x \in X} \sum_{\mu \in Y} G(x; \mu, \sigma'^2),   \text{else} \\
\end{cases}
\end{equation}

To preserve the original partial point cloud 
as perfectly as possible, we also need to calculate the KL Divergence backward to 
determine which points are the structure points of the missing parts. 
We merge these structure points with the structure points of the original partial point cloud 
to form the final output. 
In summary, we obtained a reasonable complete structure point corresponding to 
each partial point cloud through structure retrieval.

\subsubsection{Comparison with CD/EMD}

\begin{figure}[t]
\begin{center}
   \includegraphics[width=1.0\linewidth]{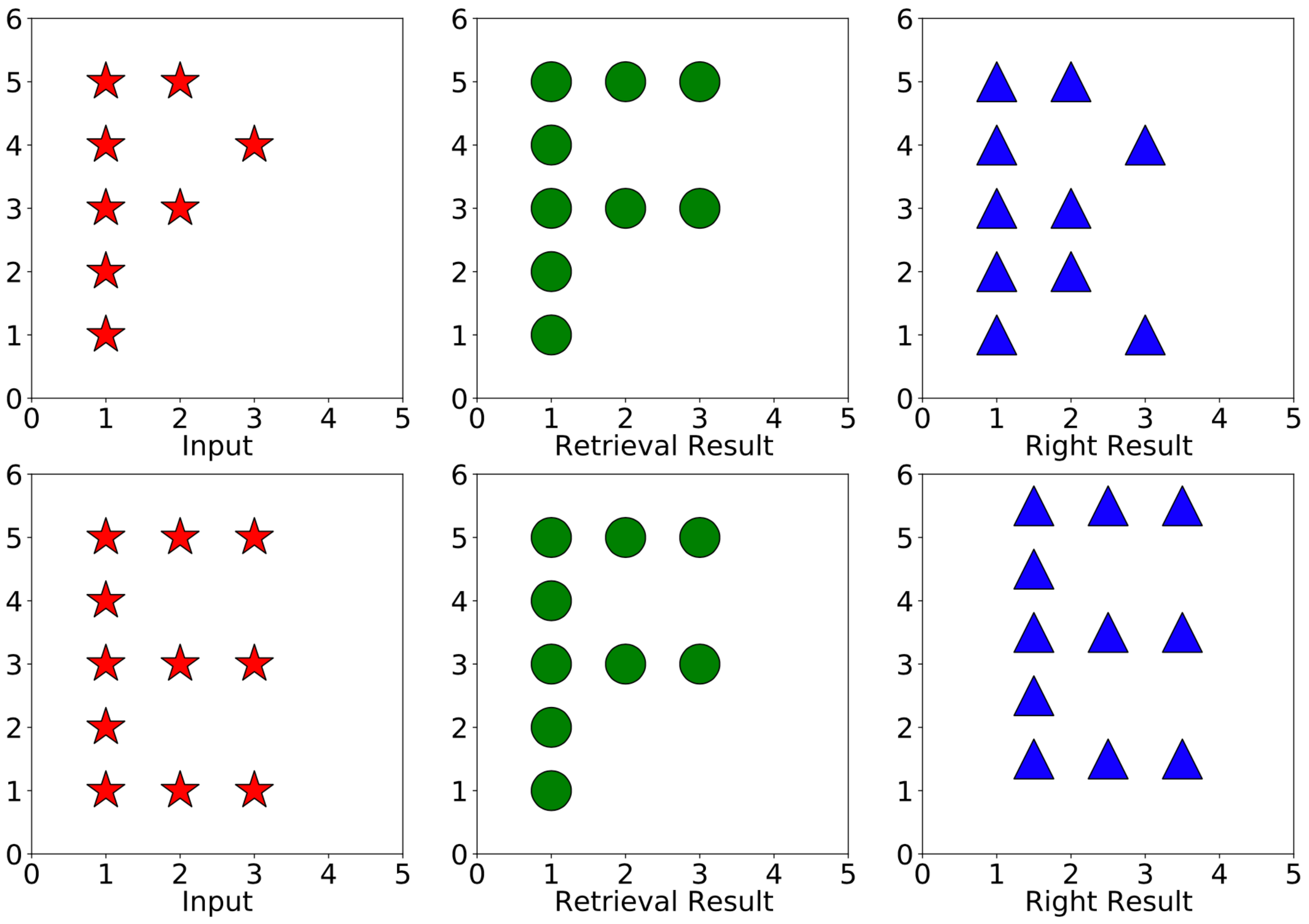}
\end{center}
   \caption{The proof of the CD (top) and  Pre\_GT's (bottom) dissatisfaction of 
   subset matching and discrete adaptation in structure retrieval.}
\label{fig:cd_emd_comparison}
\end{figure}

In the previous section, we introduce a novel measurement of the matching degree  
by using KL Divergence, 
which well satisfies the two important properties required by structure retrieval: 
subset matching and discrete adaptation. In this section, based on these two properties, 
we will explain why Chamfer Distance(CD) and Earth Mover's Distance(EMD) 
are not suitable for structure retrieval.

CD and EMD are first adopted by Fan et al.\cite{Fan_2017_CVPR} in point cloud research 
and are widely used as loss functions or 
evaluation metrics for measuring the proximity of two point clouds. Their formulas are as follows, 
where $S_1$ and $S_2$ represent two point clouds respectively: 
\begin{equation}
CD(S_1, S_2) = \frac{1}{S_1} \sum_{x \in S_1} \min_{y \in S_2} || x-y ||_2^2 + \frac{1}{S_2} \sum_{y \in S_2} \min_{x \in S_1} || y-x ||_2^2
\end{equation}
\begin{equation}
EMD(S_1, S_2) = \min_{\phi: S_1 \to S_2} \frac{1}{|S_1|} \sum_{x \in S_1} || x - \phi(x) ||_2
\end{equation}

We can find that EMD requires the same number of points in $S_1$ and $S_2$, 
which obviously does not satisfy the property of the subset matching required by structure retrieval, 
so here we focus on the analysis of CD. A specific example is shown in the top three figures of 
Figure \ref{fig:cd_emd_comparison}. 
The input is a letter $P$, and the letters to be matched are $F$ and $R$. 
We can clearly identify that $P$ is a subset of $R$. But through calculating the CD, 
we get:
\begin{equation}
CD(P, F) = \frac{25}{72} < CD(P, R) = \frac{1}{2}
\end{equation}

It means that the actual matched result is $F$, 
rather than the correct result $R$, so CD does not satisfy the property of subset matching in structure retrieval.

Furthermore, we consider that half CD, prediction to the ground truth (Pre\_GT)\cite{Gadelha_2018_ECCV, lin2018learning}, 
can well satisfy the property of subset matching. However, since the calculation of Pre\_GT is discrete, 
it cannot satisfy the property of discrete adaptation. A specific example is shown in the bottom three figures 
of Figure \ref{fig:cd_emd_comparison}. 
The input is a letter $E$, and the letters to be matched are $F$ and a slightly shifted $E$. 
We can clearly identify that $E$ should match $E$. But through calculating the Pre\_GT, 
we get:
\begin{equation}
Pre\_GT(E, F) = \frac{5}{11} < Pre\_GT(E, E') = \frac{1}{2}
\end{equation}

It means the actual matched result is $F$, rather than the correct result the slightly shifted $E$, 
so Pre\_GT does not satisfy the property of discrete adaptation in structure retrieval.

\subsection{Point Cloud Upsampling}

With the matched structure point clouds, 
we further modify the PCN Decoder and adapt it to point cloud upsampling. Specifically, 
the spatial coordinate of each structure point is concatenated with the feature embedding obtained by 
the PCN encoder and a shared-MLP is used to generate M offsets (M is the upsampling rate). 
Then we accumulate these offsets to the original structure points and get the complete point cloud. 
Finally, following MSN\cite{liu2019morphing}, 
we use EMD as the loss function. Due to the retrieved structure information, 
the upsampling method guarantees the authenticity of the completion results.

\begin{figure*}[h]
\begin{center}
    \includegraphics[width=1.0\linewidth]{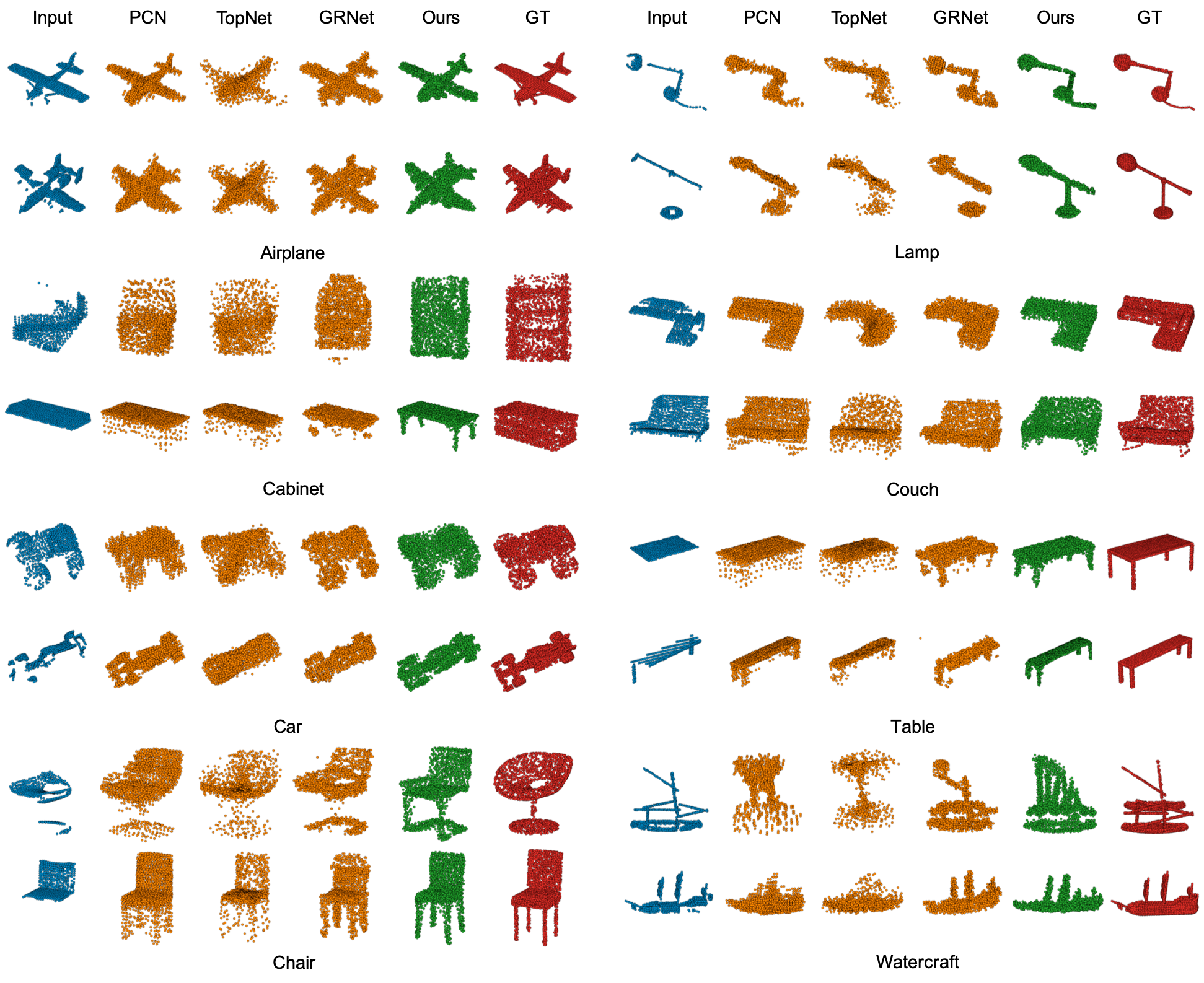}
\end{center}
    \caption{Qualitative completion results on the validation set of Completion3D dataset. 
    Since the guarantee of reasonable structure point clouds in advance, 
    we can find that the results of our method are more authentic.}
\label{fig:methods_comparison}
\end{figure*}

\section{Experiments}

\subsection{Datasets and Implementation Details}

We evaluate our experiments on the Completion3D\cite{topnet2019} dataset 
generated from the ShapeNet dataset\cite{Chang:2015:SAI}. 
It includes 30974 models and 8 categories: 
airplane, cabinet, car, chair, lamp, couch, table, watercraft. 
We pre-train our missing rate prediction model for 200 epochs, 
and then execute k-means clustering in multiple processes to get the structure point clouds. 
We set the number of clusters $K$ to 64 in our experiments. 
The structure retrieval runs on an Nvidia GPU for about 0.04s per shape. 
With the retrieved structure, we train our upsampling model on an Nvidia GPU
for 200 epochs with a batch size of 32. 
Adam is used as the optimizer and the initial learning rate is 1e-3. 

\subsection{Comparison with Existing Methods}

\begin{figure*}[h]
    \begin{center}
    \includegraphics[width=1.0\linewidth]{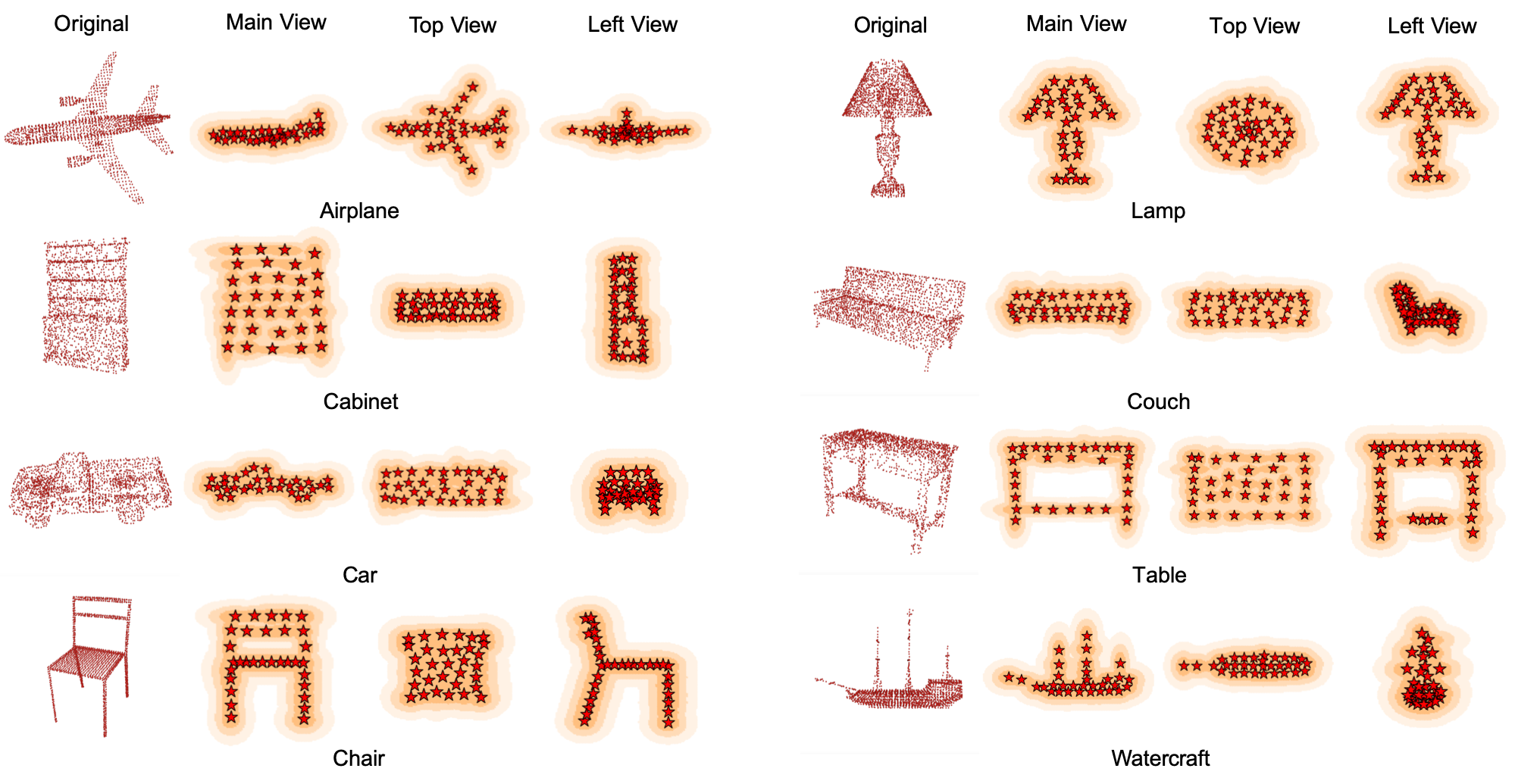}
\end{center}
    \caption{The three views of structure distribution in each category. 
    The star represents the structure point and the color from deep to shallow 
    corresponds to the value of distribution from high to low. We can find that a small number of points combined with 
    the distribution well represent the structure information.}
\label{fig:distribution_rationality}
\end{figure*}

\begin{table*}[h]
\centering
\caption{Results of Chamfer Distance ($10^{-4}$) on the test set of Completion3D dataset.}
\label{tab:cd-comparison}
\begin{tabular}{c|*8{c}|c}
    \toprule
    Methods & Airplane & Cabinet & Car & Chair & Lamp & Couch & Table & Watercraft & Overall \\
    \midrule
    PCN\cite{yuan2018pcn} & 9.79 & 22.70 & 12.43 & 25.14 & 22.72 & 20.26 & 20.27 & 11.73 & 18.22 \\
    TopNet\cite{topnet2019} & 7.32 & 18.77 & 12.88 & 19.82 & 14.60 & 16.29 & 14.89 & 8.82 & 14.25 \\
    GRNet\cite{xie2020grnet} & \textbf{6.13} & \textbf{16.90} & \textbf{8.27} & \textbf{12.23} & \textbf{10.22} & \textbf{14.93} & \textbf{10.08} & \textbf{5.86} & \textbf{10.64} \\
    \midrule
    SRPCN(Ours) & 16.06 & 35.02 & 12.6 & 42.47 & 46.43 & 26.67 & 35.4 & 13.06 & 28.67 \\
    \bottomrule
\end{tabular}
\end{table*}

We make quantitative and qualitative comparisons with PCN\cite{yuan2018pcn}, TopNet\cite{topnet2019} 
and the current best method GRNet\cite{xie2020grnet} on the Completion3D dataset. 
Since our method relies on the intermediate representation of reasonable structure point clouds, 
inaccurate but reasonable retrieval results might cause the output to be different from the ground truth. 
As shown in Table \ref{tab:cd-comparison}, our method performs poorly on the metric of Chamfer Distance. 
The data of other methods in Table \ref{tab:cd-comparison} are from Completion3D benchmark. 
But it needs to be pointed out that even if some retrieval results are inaccurate, 
the authenticity of the output will not be affected, in other words, 
reasonable objects different from the ground truth will output. 
As shown in Figure \ref{fig:methods_comparison}, our method performs well on most objects. 
Although we may match to a point cloud that is different from the ground truth, 
the result is actually quite reasonable for the input, such as the second example of the cabinet, 
the first example of the chair, and the first example of the watercraft in Figure \ref{fig:methods_comparison}. 
Compared with other methods that may complete ambiguous shapes, 
at least the results of our method are more authentic. 
Overall, the qualitative completion results confirm the necessity of reasonable structures, 
which contributes to the authenticity of final outputs.

\subsection{Distribution Rationality}

In this section, we visually show the rationality of the new proposed representation of point clouds: 
structure distribution. It includes two parts: structure points and distribution. 
As shown in Figure \ref{fig:distribution_rationality}, we give three views of some objects. 
The color from deep to shallow corresponds to the value of distribution from high to low. 
We hope that the distribution will become an envelope of the object, which can 
reduce the instability of structure points and retrain more details of the original point cloud. 
It can be found that the distributions indeed meet our expectation in Figure \ref{fig:distribution_rationality}.

\subsection{Generalization Experiments}

\begin{figure*}[h]
\begin{center}
    \includegraphics[width=1.0\linewidth]{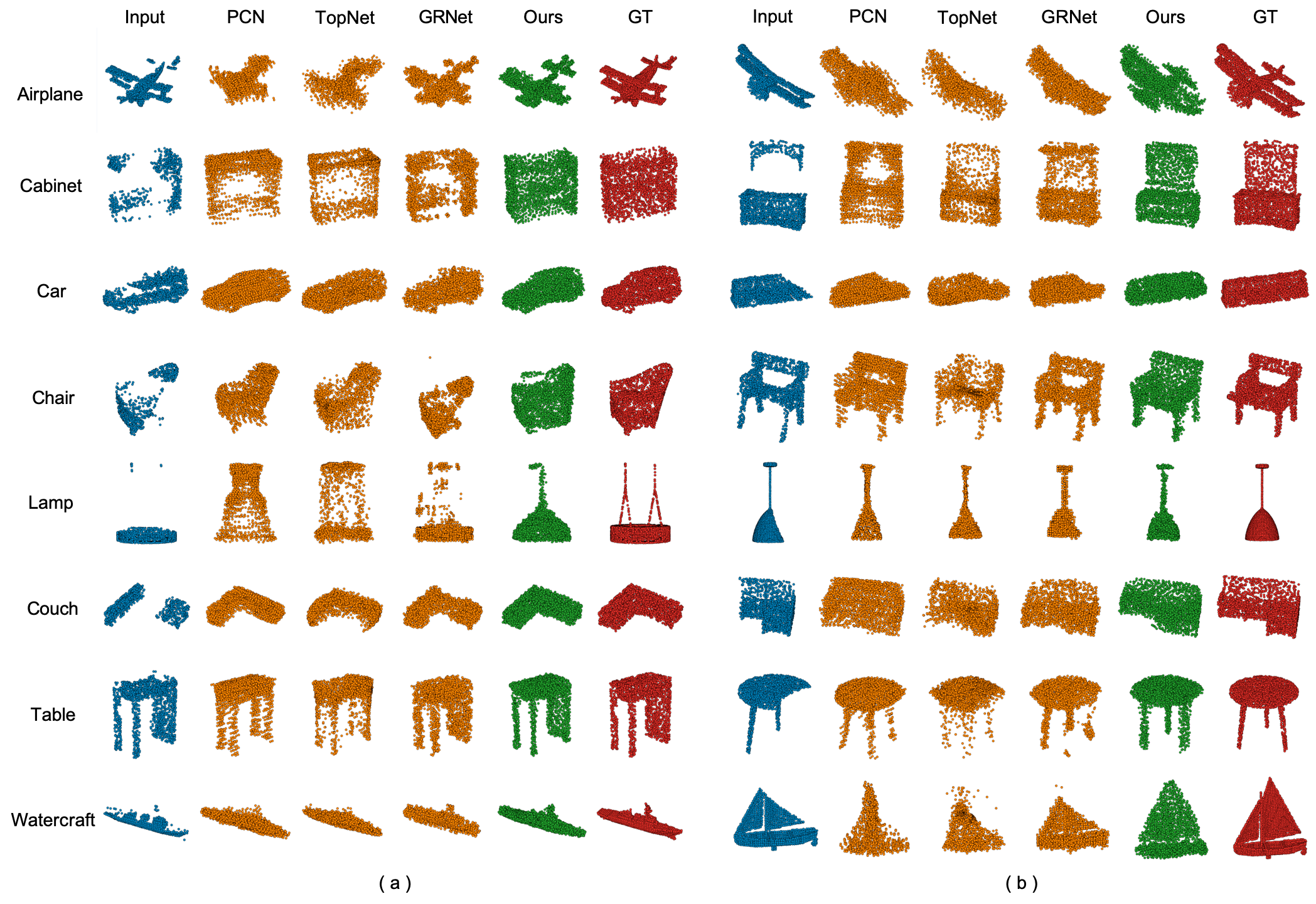}
\end{center}
    \caption{Qualitative completion results on two new missing patterns. (a) We use missing parts
    in the original dataset as input. (b) The same missing pattern as to what in PF-Net\cite{huang2020pfnet}. 
    The two new missing patterns show the generalization ability of our methods.}

\label{fig:generalization_experiments}
\end{figure*}

In this section, we prove that our method is robust to different missing patterns, 
i.e., it has a strong generalization ability. 
We conduct our experiments on two new missing patterns in the validation set of the Completion3D dataset: 
1. We use missing parts in the original dataset as input. 
2. The same as to what in PF-Net\cite{huang2020pfnet}, i.e., we remove some points 
within a certain radius from complete point clouds in a view-point. 
For the first pattern, because it is similar to the original incompleteness 
caused by back-projecting 2.5D depth images into 3D, 
the differences in the completion results of each method are insignificant. 
But overall, as shown in (a) of Figure \ref{fig:generalization_experiments}, 
the completed shape of our method is much better and more authentic. 
For the second pattern, since it is quite different from 
what in the original training set, 
the existing methods sometimes are not able to do completion, like the completion of the 
airplane, the cabinet, the car, and the watercraft in (b) of Figure \ref{fig:generalization_experiments}.

\section{Discussion}

The prediction of the missing rate is a difficult problem. 
For example, when only the back of a chair is left, it is hard to accurately predict 
due to the various size of the chair surface and legs. 
The structure points used in our retrieval method might lead to the loss of some details in the input. 
For example, we may match a relatively large square table when giving the structure points of a round tabletop. 

Some recent papers \cite{chen2020pcl2pcl, wu_2020_ECCV} point out that shape completion 
does not necessarily have paired data in reality, 
and the irreversible incompleteness may lead to the diversity of results. 
Regardless of whether the completion results are diverse or not, authenticity must be guaranteed. 
Existing evaluation metrics are not suitable for evaluating authenticity, which can be studied in the future.

\section{Conclution}

We propose a novel point cloud completion method based on structure retrieval. 
Based on the method of PCN, it adds an intermediate representation of reasonable structure point clouds 
to make the output more authentic. Considering the diversity of missing patterns, 
one input may correspond to multiple different reasonable outputs. 
We design a retrieval method to solve this problem. Specifically, it includes three steps: 
K-means clustering is adopted to get structure points; each structure point forms an ellipsoidal Gaussian distribution; 
these distributions are merged to form an overall distribution and KL Divergence is used to guide the retrieval. 
Our method does well on the situation of reversible incompleteness 
and can generate more reasonable results for the situation of irreversible incompleteness, 
while other methods may give ambiguous results. In addition, the generalization experiments also show that 
our method is more robust to different missing patterns.

{\small
\bibliographystyle{ieee_fullname}
\bibliography{egbib}
}

\end{document}